\documentclass[letterpaper, 10 pt, conference]{ieeeconf}  % Comment this line out if you need a4paper
\IEEEoverridecommandlockouts                              % This command is only needed if 
\usepackage{mathptmx} % assumes new font selection scheme installed
\usepackage{amsmath} % assumes amsmath package installed
\usepackage{amssymb}  % assumes amsmath package installed
\usepackage{xcolor}
\usepackage{booktabs}
\usepackage{censor}

\usepackage[most]{tcolorbox}

\newcommand{\cmark}{\checkmark}
\newcommand{\xmark}{$\times$}

\title{\LARGE \bf
Attention-based Hierarchical Variational Information Bottleneck for Robust Multi-Agent Communication under Variable Bandwidth
}

\author{
    Lukas Koch Vindbjerg$^{1}$, Qi Zhang$^{1}$, Yury Brodskiy$^{3}$ and Lukas Esterle$^{1,2}$%
    \thanks{*This work was supported by the ITEA ADVISOR project funded by the Innovation Fund Denmark under grant number 3156-00010B}.%
    \thanks{$^{1}$Department of Electrical and Computer Engineering, Aarhus University, Aarhus, Denmark
        {\tt\small lkvukas.vindbjerg, qz, lukas.esterle@ece.au.dk}}%
    \thanks{$^{2}$DIGIT, Aarhus University, Aarhus, Denmark}%
    \thanks{$^{3}$EIVA a/s, Denmark}%
}

\begin{document}
%comment next line out to hide my comments
\newcommand{\ybr}[1]{\textcolor{green}{#1}}

\maketitle
\thispagestyle{empty}
\pagestyle{empty}

%%%%%%%%%%%%%%%%%%%%%%%%%%%%%%%%%%%%%%%%%%%%%%%%%%%%%%%%%%%%%%%%%%%%%%%%%%%%%%%%

\begin{abstract}
Learning-based multi-agent communication under limited bandwidth does not only require deciding what to communicate, but also structuring messages so that partial transmissions remain useful. We study this problem under prefix truncation, where only the first part of each message is received. To address it, we propose \textbf{AH-VIB}, an attention-based autoregressive variational communication model that combines a variational information bottleneck (VIB) with sequential message generation and a hierarchical robustness loss. We evaluate AH-VIB on a custom cooperative object-inspection and occupancy-mapping task, where agents equipped with a limited field-of-view sensor coordinate to scan inspection objects in an occupancy-grid world, under variable and fixed bandwidth conditions, and compare it against MADDPG, CommNet, a flat VIB baseline, and an autoregressive MLP ablation. AH-VIB achieves competitive mean return while improving performance reliability under the most constrained bandwidth conditions.

These results indicate that AH-VIB improves the reliability and graceful degradation of learned communication under bandwidth constraints.
\end{abstract}

\section{Introduction}
\label{sec:intro}

Multi-robot systems increasingly support tasks such as infrastructure
inspection, environmental monitoring, and operation in hazardous
environments~\cite{dorigo2020reflections}. Effective coordination in such
settings depends on communication, yet available bandwidth can vary with
range, interference, and environmental occlusion~\cite{achord,
gielis2022review}. Robots must therefore coordinate even when only part of a
teammate's intended message is received.

\begin{figure}[t]
    \centering
    \begin{minipage}[b]{0.49\columnwidth}
        \centering
        \includegraphics[width=\textwidth]{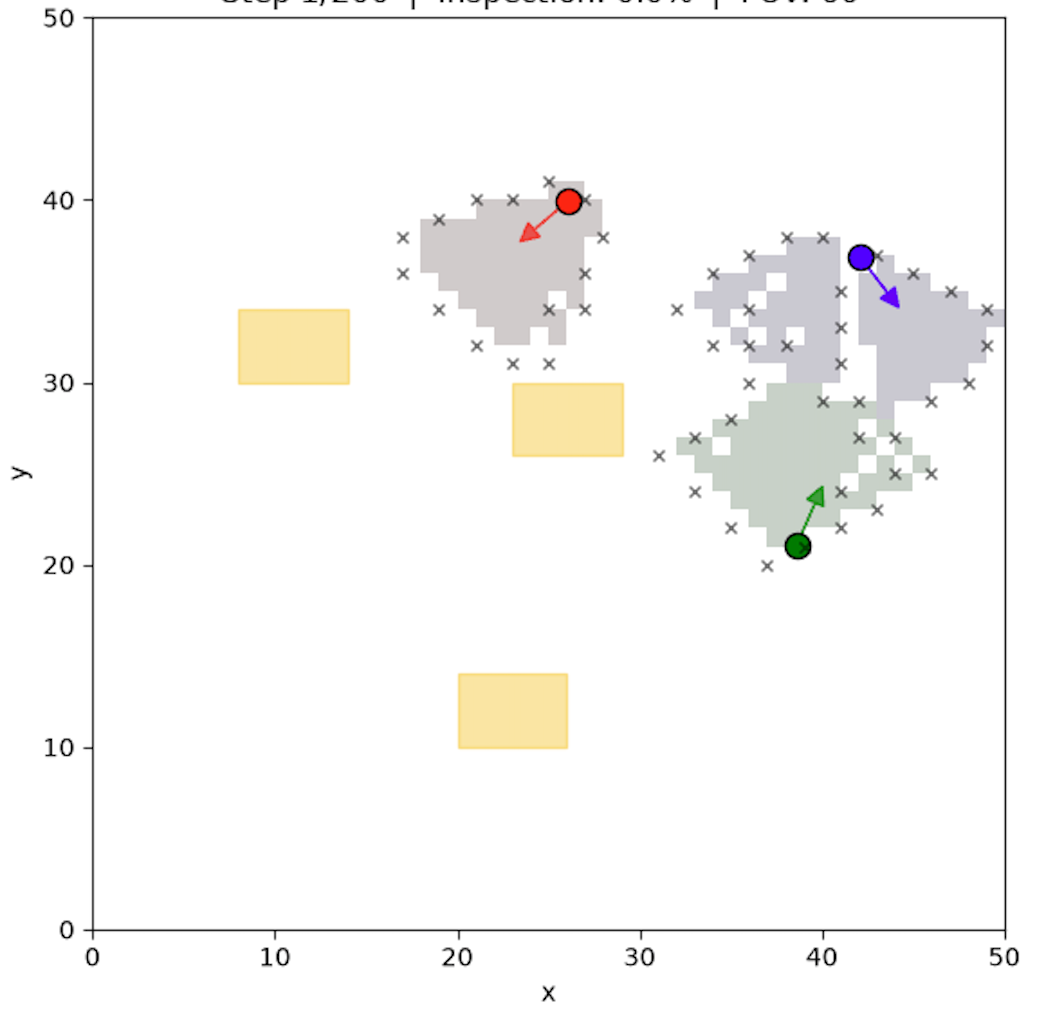}
        \centerline{\small (a) Step 1}
    \end{minipage}
    \hfill
    \begin{minipage}[b]{0.478\columnwidth}
        \centering
        \includegraphics[width=\textwidth]{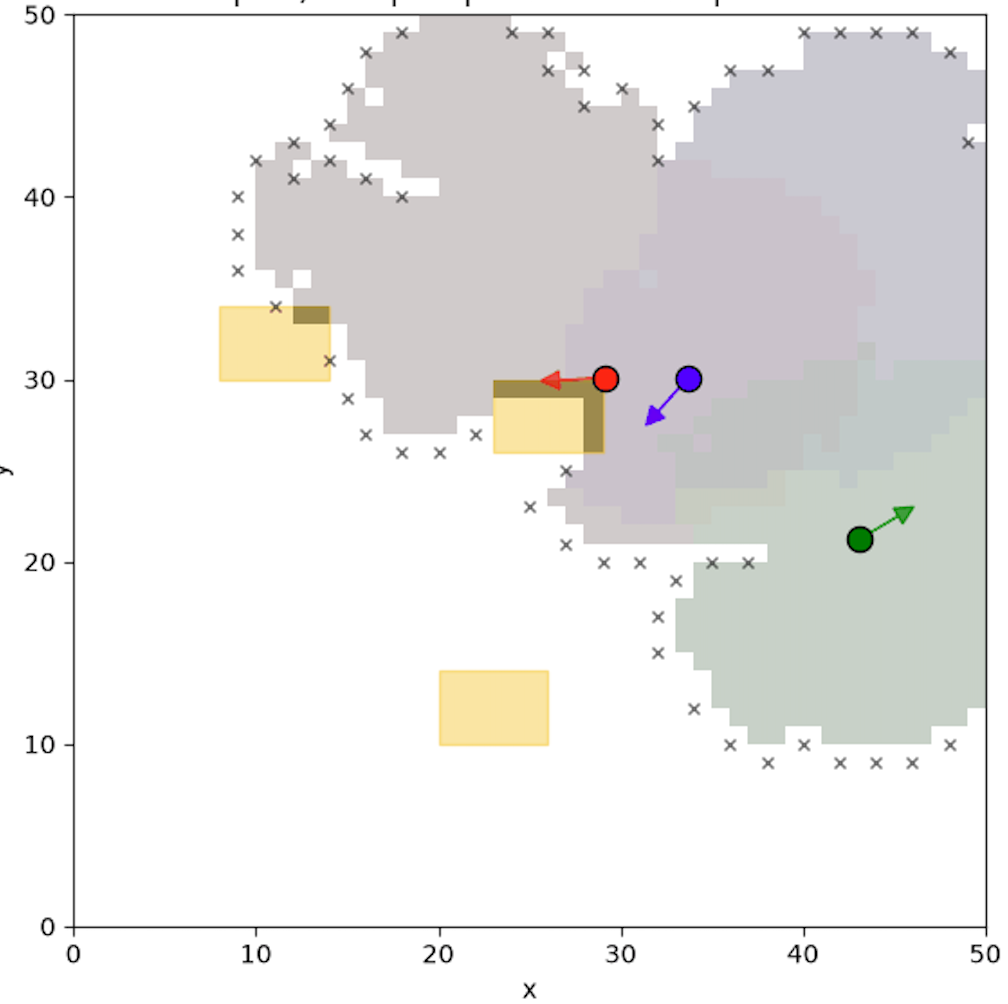}
        \centerline{\small (b) Step 50}
    \end{minipage}
    \caption{
    Cooperative object inspection under partial observability. Agents with
    limited fields of view progressively reveal free space and inspection-object
    surfaces while coordinating their coverage.
    }
    \label{fig:slam_env}
\end{figure}

We consider cooperative object inspection as a representative use case.
Agents with limited fields of view must exchange compact information about their observations, relative state, and remaining viewpoints to avoid redundant coverage. As illustrated in Figure~\ref{fig:slam_env}, a useful communication protocol should support complementary inspection even when individual agents cannot observe the complete environment. When the available communication budget changes, the received portion of a message should provide a useful coarse description rather than an arbitrary fragment.

Multi-agent reinforcement learning (MARL) has produced many learned communication protocols~\cite{commnet, das2019tarmac, zhang2021marl}, while information-theoretic methods such as IMAC~\cite{wang2020imac} use the
Variational Information Bottleneck (VIB)~\cite{vib} to learn compact messages.
However, many learned communication methods assume that messages are either fully delivered or entirely lost~\cite{gielis2022review}. We instead model variable bandwidth through \emph{prefix truncation}. Only the first $k$ dimensions of a message are received before transmission ends. This is a controlled abstraction of progressive communication degradation; practical channels may additionally involve packet loss, delay, asymmetry, and retransmission, which are outside the present scope.

Prefix truncation poses an ordering problem: \emph{what should an agent say first}? For a  message to be useful, it should place task-relevant information in early dimensions, so that short prefixes remain useful and later dimensions provide refinement. We call this property \emph{progressive information encoding}.
Thus, our objective is not merely to compress communication, but to learn ordered, task-oriented message representations.

To achieve this, we propose \textbf{AH-VIB} (Attention-based Hierarchical Variational
Information Bottleneck), which combines two complementary components. First, an autoregressive attention-based encoder generates each message dimension conditioned on the preceding prefix, making truncated prefixes structurally consistent. Second, a hierarchical robustness loss penalizes reductions in the centralized critic's estimate of team return when messages are truncated during training. Together, these components encourage messages that remain useful under changing bandwidth.

We evaluate AH-VIB on a custom cooperative object-inspection and occupancy-mapping task, in which agents with limited fields of view coordinate to scan relevant inspection objects. We compare our communication approach, used to enable the coordination among agents, against MADDPG~\cite{lowe2017maddpg}, CommNet~\cite{commnet}, a flat VIB baseline~\cite{wang2020imac}, and an autoregressive MLP ablation (HVIB). AH-VIB achieves its strongest performance at the most constrained bandwidths and provides more reliable behavior than the VIB and MADDPG baselines when only short message prefixes are available.

Our contributions can be summarized as follows:
\begin{itemize}
    \item We formulate variable-bandwidth MARL as prefix truncation of learned message vectors, as a controlled abstraction of progressively available  communication~\cite{achord, gielis2022review}.
    
    \item We propose an autoregressive attention-based message encoder that  imposes a causal ordering over message dimensions, combined with a  hierarchical robustness objective that encourages useful message prefixes.
    
    \item We evaluate AH-VIB in a cooperative object-inspection task under fixed and variable bandwidth, including return distributions and representation-level analysis of learned message usage.
\end{itemize}
\section{Related Work}
\label{sec:related}

% Our work will both look at learned multi-agent communication and bandwidth-constrained robotics. Prior work in MARL communication has largely focused on \emph{what} to communicate and \emph{to whom}, by developing advanced protocols, attention-based targeting, and information-theoretic compression. The variable-bandwidth topic, where a message is only partially received, has received far less attention. We choose to explore a different question by looking at, not how to select or compress a message, but how to \emph{order} its content so that any truncated message, however short, constitutes a coherent and maximally useful partial representation under the given constraints.

Prior MARL communication work has primarily studied what to communicate and to whom, through learned protocols, recipient targeting, and information-theoretic compression. We instead study how information should be ordered within a message when only a prefix may be received.

\subsection{Learned Communication in MARL}
CommNet~\cite{commnet} was one of the earliest demonstrations of fully differentiable multi-agent communication. CommNet specifically proposed continuous communication via fixed mean pooling of hidden states over $K$ rounds, learning both a policy and a communication protocol end-to-end. While effective on cooperative tasks, CommNet treats all agents' messages symmetrically and provides no mechanism for selective or bandwidth-adaptive encoding.

MADDPG~\cite{lowe2017maddpg} popularized the Centralized Training with Decentralized Execution (CTDE) paradigm, where a centralized critic evaluates during training with decentralized execution of agents. This enables stable multi-agent policy gradient updates.
TarMAC~\cite{das2019tarmac} extended learned communication with a signature-based soft attention mechanism, allowing agents to learn \emph{what} to communicate and \emph{to whom}. 
Our work is complementary. Where TarMAC addresses targeting (who receives a message), we address content encoding under bandwidth constraints (what survives when the channel is limited).

\subsection{Information-Theoretic Communication}

The Variational Information Bottleneck (VIB)~\cite{vib}provides a principled framework for learning compressed representations that retain only task-relevant information, originally developed outside the multi-agent setting. IMAC~\cite{wang2020imac} adapted this framework to multi-agent communication, applying VIB to message encoding and using KL regularization to learn compact representations that balance task performance against communication cost under bandwidth constraints. Our work builds directly on IMAC's VIB formulation but identifies its key limitation under prefix truncation. A flat VIB encoder produces all message dimensions simultaneously from a shared hidden state, imposing no ordering across dimensions. 
AH-VIB instead uses an autoregressive factorization that makes each prefix
causally consistent, while the hierarchical robustness loss encourages
prefixes to retain task utility.

\subsection{Bandwidth-Constrained Multi-Robot Communication}

Outside the MARL framework, bandwidth constraints in real multi-robot systems have been studied extensively in the robotics literature.
ACHORD~\cite{achord} demonstrates that real multi-robot deployments in non-line-of-sight environments experience intermittent connectivity, with robots accumulating data backlogs in on-board queues that must be transmitted when a connectivity window is available. This stands in stark contrast to the full-delivery assumption of most learned communication methods.

% \subsection{Hierarchical and Progressive Message Encoding}
% The concept of bandwidth-ordered message encoding draws inspiration from the long-established success of progressive refinement in source coding. In image compression~\cite{shapiro2002embedded}, scalable video, and audio coding, representations are structured so that early portions constitute a coarse but complete description, with successive portions adding fidelity. We adopt this concept and impose it on a learned latent space to take advantage of this property of graceful degradation under any transmission cutoff.

\subsection{Hierarchical and Progressive Message Encoding }
Progressive or embedded source coding produces representations whose initial bits provide a useful coarse description and whose later bits refine it.
Embedded zerotree wavelet coding~\cite{shapiro2002embedded}, for example, orders coded information by importance so that encoding or decoding can stop at an arbitrary rate. Related ideas appear in learned ordered representations: nested dropout~\cite{rippel2014learning} and Matryoshka
Representation Learning~\cite{kusupati2022matryoshka} train nested latent prefixes for adaptive compression or variable representation budgets. We adapt this principle to learned multi-agent communication. The received prefix models the portion of a message that survives a variable-bandwidth channel, and AH-VIB combines autoregressive generation with a hierarchical robustness loss to retain task utility under truncation.

\section{System Model}
\label{sec:background}

We consider a cooperative Multi-Agent Reinforcement Learning (MARL) setup where $n$ agents coordinate their actions to achieve a shared objective under partial observability. Each agent relies both on its local observations and also on explicit communication with other agents~\cite{Esterle2022loosening}.

\subsection{Dec-POMDP Formulation}

We formally describe the problem as a communication-augmented Decentralized Partially Observable Markov Decision Process (Dec-POMDP), defined by the tuple:

\[
\langle N,\, S,\, \{\hat{A}_i\}_{i=1}^n,\, \{O_i\}_{i=1}^n,\, M,\, T,\, R,\, \gamma \rangle~.
\]

Here, $N = \{1, \ldots, n\}$ is the set of agents and $S$ is the global state space of the environment. 
Each agent $i$ has a composite action space $\hat{A}_i = A_i \times M \subset \mathbb{R}^{d_i^a+d_m}$, so that a composite action is the pair $(a_i,m_i)$.
Here, $a_i \in A_i \subset \mathbb{R}^{d_i^a}$ is the physical environment action and $m_i \in M \subset \mathbb{R}^{d_m}$ is the message, where $d_i^a$ and $d_m$ denote the physical-action and message dimensionalities, respectively.
Agent $i$ also has a local observation space $o_i\in O_i \subset \mathbb{R}^{d_i^o}$. The shared message space $M$ is common to all agents. Each agent $i$ sends $m_i$ and receives messages from all other agents $\{m_j\}_{j \in N_{-i}}$, where $N_{-i} = N \setminus \{i\}$. 
The state transition function $T: S \times A_1 \times \cdots \times A_n \rightarrow \Delta(S)$ maps joint physical actions to a probability distribution over next states. 
Messages do not directly affect the world state. Finally, $R: S \times A_1 \times \cdots \times A_n \times S \rightarrow \mathbb{R}$ is the shared team reward function, and $\gamma \in [0, 1)$ is the discount factor.

At each timestep $t$, agent $i$ receives a local observation $o_i^t$ and the tuple of messages sent by all other agents at the previous timestep,
\begin{equation}
    \mathbf{m}_{-i}^{\,t-1} = \{m_j^{t-1}\}_{j \in N_{-i}},
\qquad N_{-i} = N \setminus \{i\}.
\end{equation}

\begin{figure}[t]
    \centering
    \vspace{2mm}
    \includegraphics[width=0.7\columnwidth, trim=0cm 0.75cm 0cm 0.55cm, clip] % trim = left bottom right top
    {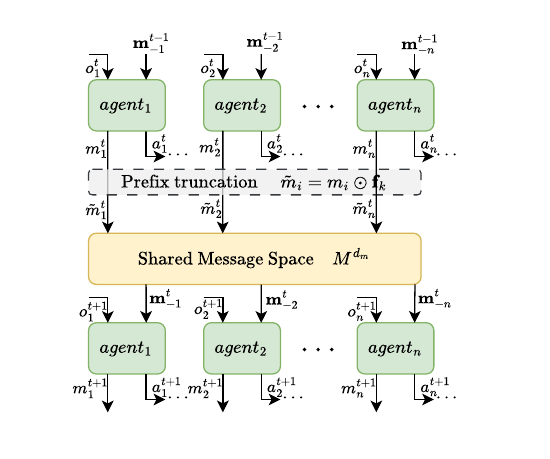}
        \caption{
        Execution flow of the multi-agent system at timestep $t$. Agent $i$ receives local observation $o_i^t$ and the message set $\mathbf{m}_{-i}^{\,t-1} = \{m_j^{t-1}\}_{j \in N_{-i}}$ sent by the other agents at the previous timestep. From these inputs, it produces a physical action $a_i^t$ and a message $m_i^t$. Before delivery, $m_i^t$ is truncated by the channel, retaining only the first $k$ dimensions, and is then broadcast to the other agents for use at timestep $t+1$.
}

    \label{fig:marl_flow}
\end{figure}

At episode start, all messages are initialized to zero. 
The agent then selects a composite action $\hat{a}_i^t = (a_i^t, m_i^t) \in \hat{A}_i$ according to its policy $\pi_\theta$ as, 
\begin{equation}
(a_i^t, m_i^t) = \pi_\theta^i\!\left(o_i^t,\; \mathbf{m}_{-i}^{\,t-1}\right).
\end{equation}
% The agent then selects a composite action
% $(a_i^t, m_i^t) \in \hat{A}_i$ according to
% \begin{equation}
% \pi_\theta^i : O_i \times M^{\times (n-1)} \rightarrow A_i \times M.
% \end{equation}
Figure~\ref{fig:marl_flow} illustrates this execution flow, including the one-step delay between message transmission and reception.

The objective is to learn a joint policy $\boldsymbol{\pi} = (\pi_1, \ldots,
\pi_n)$ that maximizes the expected discounted return:

\begin{equation}
  J(\boldsymbol{\pi}) = \mathbb{E}_{\boldsymbol{\pi}} \left[ \sum_{t=0}^{\infty}
\gamma^t r^t \right]  ~,
\end{equation}

where $r^t = r_i^t$ for all $i \in N$ is the shared team reward emitted by the environment at timestep $t$. All agents receive an identical scalar reward, reflecting the fully cooperative nature of the task.

\subsection{Centralized Training with Decentralized Execution (CTDE)}
We use the CTDE paradigm~\cite{lowe2017maddpg}. During training, a single centralized critic receives the pooled observations and contexts of all agents as input:
\begin{equation}
 Q(\mathbf{s},\, \mathbf{a}) = Q\!\left(
[o_1 \| c_1,\; \ldots,\; o_n \| c_n],\;
[a_1 \| m_1,\; \ldots,\; a_n \| m_n]
\right)   ~,
\end{equation}
where $\mathbf{s} = [o_1 \| c_1, \ldots, o_n \| c_n]$ concatenates each agent's local observation and aggregated context, and $\mathbf{a} = [a_1 \| m_1, \ldots, a_n \| m_n]$ concatenates all agents' environment actions and messages. 
The critic is not used during execution time, where each agent operates using only its local observation $o_i$ and context $c_i$, making the operation fully decentralized.

\subsection{Communication Model}
\label{sec:comm_model}

We model bandwidth-limited communication as \textbf{prefix truncation} of the message vector. Only the first $k$ dimensions of a learned message are delivered, while the remaining dimensions are not available to the receiving agents.

Formally, given a message $m_i \in \mathbb{R}^{d_m}$ and prefix length $k \in \{1, \ldots, d_m\}$, the received message $\tilde{m}_i$ is
\begin{equation}
\tilde{m}_{i,d} =
\begin{cases}
m_{i,d}, & d \leq k, \\
0, & d > k,
\end{cases}
\qquad d = 1, \ldots, d_m,
\end{equation}
where $d$ denotes the index in the message vector.

The same $k$ is applied symmetrically to all agents. We define the bandwidth fraction as $\rho = k/d_m \in (0,1]$. At the start of each episode, $\rho$ is sampled from a Beta distribution, with $k \geq 1$ enforced by the episode-level sampler (dynamic within-episode bandwidth schedules used at evaluation may reach $k=0$). 
This formulation gives a structural requirement on the message encoder. The earliest dimensions should carry the most task-relevant information, since they are the most likely to survive at low bandwidth. We therefore study how agents should order information across message dimensions so that any received prefix remains useful for coordination, rather than treating the channel as either fully available or completely unavailable.
\section{Attention-based Hierarchical Variational Information Bottleneck}
\label{sec:approach}

Agents must learn to encode messages such that the first dimensions contain the most task-relevant information, since later dimensions may be truncated and lost entirely. Here, we present our approach to achieving this behavior.  The core contribution is two-fold: (i) an \textbf{autoregressive, attention-based message encoder} that constructs each message dimension conditioned on all previous ones, creating a natural information priority ordering. And (ii) a \textbf{hierarchical robustness loss} that actively penalizes degradation in team performance under prefix truncation during training. Together, these make up the \textbf{Attention-based Hierarchical Variational Information Bottleneck} (AH-VIB) algorithm.

The autoregressive construction provides a structural guarantee as a received prefix is generated according to the same causal factorization as the corresponding prefix of the full message. It does not, by itself, guarantee that short prefixes preserve high task value. That property is encouraged separately by the hierarchical robustness loss. Thus, the two components address different failure modes. Autoregressive generation makes prefixes structurally coherent, whereas the hierarchical loss makes them useful for the task. This distinction is important when interpreting the ablations and the relative contributions of AH-VIB's components.

\subsection{Variational Information Bottleneck}
\label{sec:vib_foundation}

The Variational Information Bottleneck (VIB)~\cite{vib} provides a principled framework for learning compressed representations that retain only task-relevant information. Rather than outputting a deterministic representation, VIB trains a stochastic encoder that maps its input to a distribution from which a sample is drawn at each forward pass.
IMAC~\cite{wang2020imac} applies this framework to multi-agent communication by applying VIB to \emph{message} encoding such that agents learn to produce compact, informative messages under bandwidth constraints. 
We adopt this approach as the probabilistic foundation for our encoder, and extend it to address the prefix-truncation setting.

\begin{table*}[htbp]
\centering
\vspace{2mm}
\caption{Training hyperparameters used in the experiments, shared across all
algorithms unless noted.}
\label{tab:hyperparams}
\begin{tabular}{ll@{\hskip 1.3em}ll@{\hskip 1.3em}ll}
\toprule
\textbf{Hyperparameter} & \textbf{Value} &
\textbf{Hyperparameter} & \textbf{Value} &
\textbf{Hyperparameter} & \textbf{Value} \\
\midrule

\multicolumn{2}{l}{\textit{Environment \& Training}} &
\multicolumn{2}{l}{\textit{Optimisation}} &
\multicolumn{2}{l}{\textit{Exploration Noise}} \\
\midrule
Total env.\ steps               & $5 \times 10^5$      & Actor LR                    & $3 \times 10^{-4}$        & Initial noise $\sigma_0$       & $0.20$ \\
Episode length $T$              & $200$                   & Critic LR                   & $3 \times 10^{-5}$        & Decay factor                   & $0.80$ \\
Replay buffer size              & $2 \times 10^5$        & Discount $\gamma$           & $0.95$                    & Decay over (steps)             & $1 \times 10^5$ \\
Batch size                      & $256$                  & Target network $\tau$       & $0.005$                   & Min.\ noise $\sigma_{\min}$    & $0.05$ \\
Random exploration steps        & $4{,}000$              & Actor hidden layers         & $[512,\;256,\;256,\;128]$ &                                 &      \\
Update interval                 & $50$                   & Critic hidden layers        & $[512,\;256,\;256,\;128]$ &                                 &      \\
Gradient steps / update         & $5$                    &                             &                           &                                 &      \\
Random seeds                    & $3$  &                             &                           &                                 &      \\
% Hardware                        & RTX 5090               &                             &                           &                                 &      \\
\midrule

\multicolumn{2}{l}{\textit{Communication Channel}} &
\multicolumn{2}{l}{\textit{VIB \& Hier.\ Loss}} &
\multicolumn{2}{l}{\textit{Algorithm-Specific}} \\
\midrule
Message dim.\ $d_m$             & $32$                   & VIB coeff.\ $\beta$         & $0.001$                   & Attn.\ dim.\ $d_{\text{attn}}$ & $64$ \\
BW sampling $(\alpha,\beta)$    & $(3.0,\;7.0)$          & Hier.\ loss $\lambda$       & $0.05$                    & Attention heads (AH-VIB)       & $4$ \\
Comm regimes   & full / zero / beta & Hier.\ samples / update     & $1$                       & CommNet rounds $K$             & $3$ \\
Loss probability & $0.0$ ($1.0$ in zero comms) & $p_{\text{full}}$ (hier.\ sampler) & $0.10$           &                                 &      \\
Topology                        & broadcast              & Stop-grad.\ $Q_{\text{full}}$ & \cmark           &                                 &      \\
\bottomrule
\end{tabular}
\end{table*}

Concretely, each agent $i$ encodes its input into a mean $\mu_d$ and
standard deviation $\sigma_d$ for each message dimension $d$, from which
a sample is drawn as:
\begin{equation}
    z_d = \mu_d + \sigma_d \cdot \epsilon, \qquad \epsilon \sim \mathcal{N}(0, 1)~,
    \label{eq:reparam}
\end{equation}
giving the full message $m_i = [z_1, \ldots, z_{d_m}]$. 

The encoder is regularized by minimizing the Kullback–Leibler (KL) divergence from a standard normal prior across all dimensions and agents:
\begin{equation}
    \mathcal{L}_{\mathrm{VIB}} = \sum_{i=1}^{n} \sum_{d=1}^{d_m}
    D_{\mathrm{KL}}\!\left(\mathcal{N}(\mu_d, \sigma_d^2) \,\|\, \mathcal{N}(0, 1)\right)~.
    \label{eq:vib_loss}
\end{equation}

Traditional VIB encoders predict all $(\mu_d, \sigma_d^2)$ simultaneously from a shared hidden state, which provides no intentional structure on how information is distributed across dimensions. Consequently, the first $k$ dimensions are not necessarily optimized as a self-contained representation for every $k$. As a result, any given dimension may implicitly depend on information encoded in later dimensions, meaning a truncated prefix is not necessarily a coherent or self-sufficient representation. The order in which information is placed across message dimensions must therefore be learned.

\subsection{Autoregressive Message Encoding}
\label{sec:autoregressive}

To enforce message-order priority, we model the message dimensions autoregressively, so that each dimension can depend on those generated before it. This directly addresses the cross-dimensional dependency problem identified above. Each dimension $z_d$ is conditioned on the previously generated prefix $\mathbf{z}_{<d} = (z_1,\ldots,z_{d-1})$:
\begin{equation}
    q(z_d \mid \mathbf{z}_{<d},\, o_i, c_i), \qquad d = 1, \ldots, d_m .
    \label{eq:autoregressive}
\end{equation}
Since no dimension conditions on any later $z_{\hat d}$ with $\hat d > d$, any truncated prefix remains self-consistent by construction.

We implement this using a multi-head attention mechanism. At each iteration over $d$, the encoder hidden state $h_i$ is projected into attention space and combined with a position embedding to form the query, while the key-value context grows with each previously sampled dimension:
\begin{equation}
    \mathbf{q}_d = W_{\text{proj}}\, h_i + \text{pos}(d) ~,
    \label{eq:query}
\end{equation}
\begin{equation}
    \mathrm{KV}_d = \bigl[\,W_{KV}\, h_i,\;\; W_z\, z_1,\;\;
    \ldots,\;\; W_z\, z_{d-1}\,\bigr] ~.
    \label{eq:kv}
\end{equation}
The attention output is concatenated with $h_i$ and passed through a small head network to produce $(\mu_d, \sigma_d^2)$, from which $z_d$ is sampled and bounded:
\begin{equation}
    z_d = \mu_d + \sigma_d \cdot \epsilon, \quad \epsilon \sim \mathcal{N}(0,1),
    \qquad \tilde{z}_d = \tanh(z_d) ~.
    \label{eq:sample}
\end{equation}
Note that $W_z$ is shared across all steps to reduce the number of parameters and improve generalization. The attention mechanism allows each dimension to focus on the most relevant previously generated dimensions, rather than receiving a fixed concatenation.

\begin{figure}[t]
    \centering
    \includegraphics[width=\columnwidth, trim=0cm 0.7cm 0cm 0.7cm, clip] % trim = left bottom right top
    {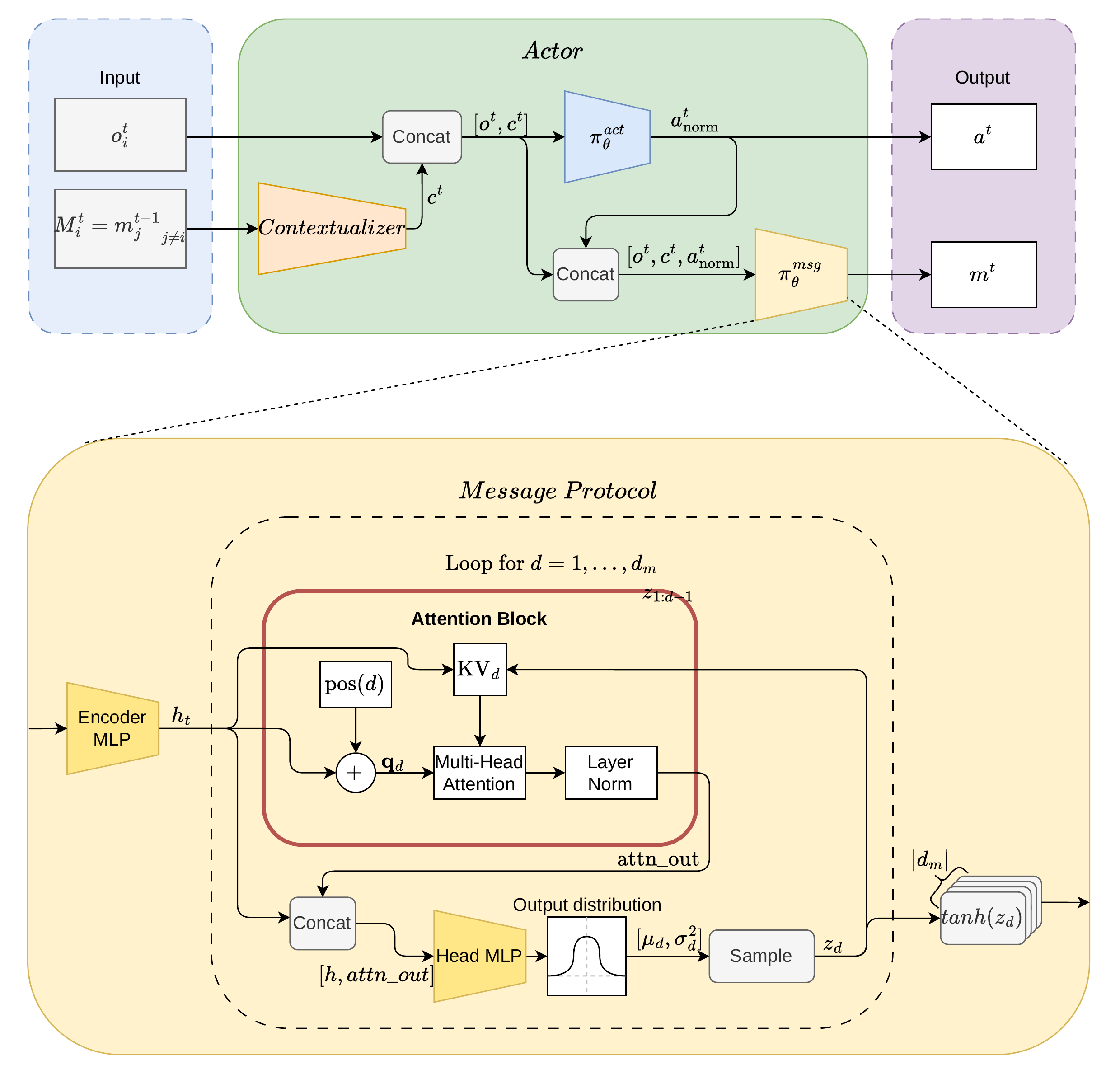}
    \vspace{4mm}
    \caption{
        AH-VIB architecture. \textit{Top:} Full agent diagram showing how observation $o_i$ and aggregated context $c_i$ are processed by the \textit{Contextualizer}-encoder MLP into hidden representation $[o^t, c^t]$, which drives both the action head $\pi^{act}_\theta$ to produce $a_i^{\text{norm}}$ and together with $a_i^{\text{norm}}$, the autoregressive message encoder to produce $m_i = [\tilde{z}_1, \ldots, \tilde{z}_{d_m}]^\top$. 
        \textit{Bottom:} Detail of the autoregressive message encoder. The dashed box is executed sequentially for $d = 1, \ldots, d_m$. At each step, the key-value context $\mathrm{KV}_d$ is constructed by stacking a projection of $h_i$ with projections of all previously sampled dimensions $\tilde{z}_{1:d-1}$, growing by one token each iteration. The query $\mathbf{q}_d$ is formed from $W_{\text{proj}}\,h_i$ and position embedding $\text{pos}(d)$. The multi-head attention output is concatenated with $h_i$ and passed through the head MLP to produce $(\mu_d, \sigma_d^2)$, from which $\tilde{z}_d$ is sampled and bounded by $\tanh$. The sampled $\tilde{z}_d$ is fed back into $\mathrm{KV}_{d+1}$ and accumulated.
    }
    \label{fig:AH-VIB_arch}
\end{figure}

The full message $m_i = [\tilde{z}_1, \ldots, \tilde{z}_{d_m}]$ is assembled sequentially and transmitted through the channel, where prefix truncation is applied before message transmission. Figure~\ref{fig:AH-VIB_arch} illustrates the complete AH-VIB architecture, with the bottom panel illustrating a single autoregressive step.

\subsection{Hierarchical Robustness Loss}
The autoregressive structure ensures representational consistency under truncation, but does not directly enforce strong task performance at reduced bandwidth.
We address this with a \textbf{hierarchical robustness loss} that directly penalizes drops in team value under prefix truncation.

During each actor update, we sample a truncation ratio $\rho$ for the robustness loss and convert it to a prefix length $k$ as:
\begin{equation}
    k = \mathrm{round}(\rho d_m) \qquad \rho \sim q(\rho),
\end{equation}
where
\begin{equation}
    q(\rho)=
    \begin{cases}
        1, & \text{with probability } p_{\text{full}}, \\[4pt] \mathrm{Beta}(\alpha,\beta), & \text{otherwise.}
    \end{cases}
\end{equation}
This truncation sampler is independent of the actual channel bandwidth applied during environment interaction. By decoupling the hierarchical loss from the environment's specific bandwidth dynamics, we encourage the encoder to learn a progressive message structure that generalizes across varying bandwidth conditions, rather than overfitting to the particular degradation schedule of the training environment.

We apply the prefix mask to the joint action vector and evaluate the
centralized critic under truncation:

\begin{equation}
   Q_k = Q\!\left(\mathbf{s},\; \tau_k(\mathbf{a})\right), \qquad Q_{\text{full}} = Q\!\left(\mathbf{s},\; \mathbf{a}\right) ~.
\end{equation}

The hierarchical loss penalizes the gap between the full and truncated team
value using a hinge:

\begin{equation}
   \mathcal{L}_{\text{hier}} =
    \mathbb{E}_{k}\!\left[\max\!\left(0,\;
    \overline{Q}_{\text{full}} - Q_k\right)\right] ~,
\end{equation}
where $\overline{Q}_{\text{full}}$ is a stop-gradient copy of $Q_{\text{full}}$, i.e., it is treated as a constant when computing gradients for $\mathcal{L}_{\text{hier}}$.
This makes the loss increase $Q_k$ toward the full-bandwidth reference, rather than reducing $Q_{\text{full}}$ to satisfy the objective.

The complete actor loss for AH-VIB is:
\begin{equation}
\mathcal{L}_{\text{actor}}
=
\mathbb{E}\!\left[
- Q_{\text{full}}
+ \beta\,\mathcal{L}_{\text{VIB}}
+ \lambda
% \max\!\left(0,\; \overline{Q}_{\text{full}} - Q_k\right)
\mathcal{L}_{\text{hier}}
\right] ~.
\end{equation}

The three terms have complementary roles: $-Q_{\text{full}}$ drives task
performance at full bandwidth; $\mathcal{L}_{\text{VIB}}$ enforces message compression; $\mathcal{L}_{\text{hier}}$ enforces graceful degradation under truncation. Thus, the two components address different failure modes. Autoregressive generation makes prefixes structurally coherent, whereas the hierarchical loss makes them useful for the task. This distinction is important when interpreting the ablations and the relative contributions of AH-VIB's components.

\section{Experiments \& Results}
\label{sec:experiments}

\subsection{Environment}

We evaluate all algorithms on a custom multi-agent \textbf{inspection SLAM} task implemented in the semcomms framework as a PettingZoo \texttt{ParallelEnv}~\cite{terry2021pettingzoo}. 
A team of $n=3$ agents explores a $50 \times 50$ occupancy grid containing three randomly placed $6 \times 4$ rectangular inspection objects. At each step, agents select a continuous 2D velocity action $u \in [-1,1]^2$ (maximum displacement: two cells); the resulting heading defines a $60^\circ$ field-of-view sensor with 16 rays and a range of eight cells.
Figure~\ref{fig:slam_env} illustrates the environment at the start of an episode and after 50 steps of exploration.

Each agent maintains its own personal occupancy grid, updated by ray-casting. Object cells occlude scans, so only surface cells can be revealed. A shared team grid (union of all personal grids) drives a team-shared reward with three additive terms:

\begin{equation}
    r^t = w_{\text{obj}} \frac{n_{\text{obj}}^t}{|\mathcal{O}|}
        + w_{\text{free}} \frac{n_{\text{free}}^t}{G^2}
        - \lambda_{\text{dup}}\, c_{\text{dup}}^t.
    \label{eq:slam_reward}
\end{equation}

Here, $n_{\text{obj}}^t$ is the number of inspection-object surface cells newly revealed in the team grid, $|\mathcal{O}|$ is the total number of object cells ($3$ objects $\times 6 \times 4 = 72$ cells), $n_{\text{free}}^t$ is the number of free cells newly revealed, $G=50$ is the grid size, $c_{\text{dup}}^t$ counts agent pairs within a proximity threshold of $3$ cells of each other, $\lambda_{\text{dup}}=0.5$, with weights $w_{\text{obj}}=1.0$ and $w_{\text{free}}=0.01$. The second term is a deliberately small exploration bonus, so discovering unoccupied geometry is non-trivially incentivized but object scanning dominates.

Each agent receives a partial observation consisting of a $17\times17$ local patch from its personal occupancy grid (normalized to $0/0.5/1$ for unknown/free/occupied) concatenated with its own normalized pose $[x/G,\, y/G,\, \cos\theta,\, \sin\theta]$; the observation dimension is $17^2+4=293$. Communication context is not part of the observation (all comm passes through the external channel module). Episodes run for $T=200$ timesteps.

The communication channel is configured as described in Section~\ref{sec:comm_model}, with synchronous broadcast, message dimension $d_m=32$, and episode-level bandwidth sampling from $\mathrm{Beta}(3.0,\,7.0)$. Prefix truncation is applied uniformly to all agents. 
For the robustness loss, we use a separate truncation sampler with the same Beta parameters and $p_{\text{full}}=0.10$. Furthermore, the training-time comm regime is ablated across three conditions: \emph{full} (forced $\rho=1$), \emph{zero} ('loss\_prob: 1' hard drop), and \emph{beta} (the default Beta-sampling truncation).
\begin{table}[tb]
\centering
\caption{Baseline design comparison.}
\label{tab:baselines}
\begin{tabular}{lcccc}
\toprule
\textbf{Algorithm} & \textbf{VIB} & \textbf{AR} &
\textbf{Hier.} & \textbf{Attn.} \\
\midrule
MADDPG                 & \xmark & \xmark & \xmark & \xmark \\
CommNet                & \xmark & \xmark & \xmark & \xmark \\
VIB                   & \cmark & \xmark & \xmark & \xmark \\
HVIB                   & \cmark & \cmark & \cmark & \xmark \\
\textbf{AH-VIB } & \cmark & \cmark & \cmark & \cmark \\
\bottomrule
\end{tabular}
\end{table}
\subsection{Baselines}

For clarity, we include both standard communication baselines and structured ablations of our own method. Alongside the full AH-VIB model, we include (i) a flat VIB-based communication model with simultaneous message generation and (ii) an autoregressive MLP-based variant, which we refer to as HVIB. These baselines provide progressively structured comparisons. The flat VIB baseline tests the effect of stochastic compression without ordered message generation. HVIB adds autoregressive generation and the hierarchical loss but replaces attention with an MLP. The comparison between HVIB and AH-VIB therefore evaluates whether attention provides benefits beyond autoregressive ordering and truncation-aware training.

We compare AH-VIB against four alternative approaches:

\begin{itemize}

\item \textbf{MADDPG}~\cite{lowe2017maddpg}: a deterministic CTDE baseline with learned continuous messages, but without a stochastic bottleneck, explicit dimensional ordering, or robustness loss.

\item \textbf{CommNet}~\cite{commnet}: a continuous communication baseline based on differentiable mean pooling over $K=3$ rounds, with no explicit bottleneck or dimensional ordering.

\item \textbf{IMAC-style VIB}~\cite{wang2020imac}: a flat variational communication baseline inspired by IMAC, where all message dimensions are generated simultaneously from a shared encoder. We implement only the communication component relevant to our setting.

\item \textbf{HVIB}: our autoregressive MLP ablation. HVIB retains the stochastic bottleneck and hierarchical robustness loss, but replaces the attention-based autoregressive message mechanism of AH-VIB with a standard per-dimension MLP-based construction.

\end{itemize}

\begin{figure}[htbp]
    \centering
    \includegraphics[width=\columnwidth, trim=0cm 0.8cm 0cm 0.8cm, clip]{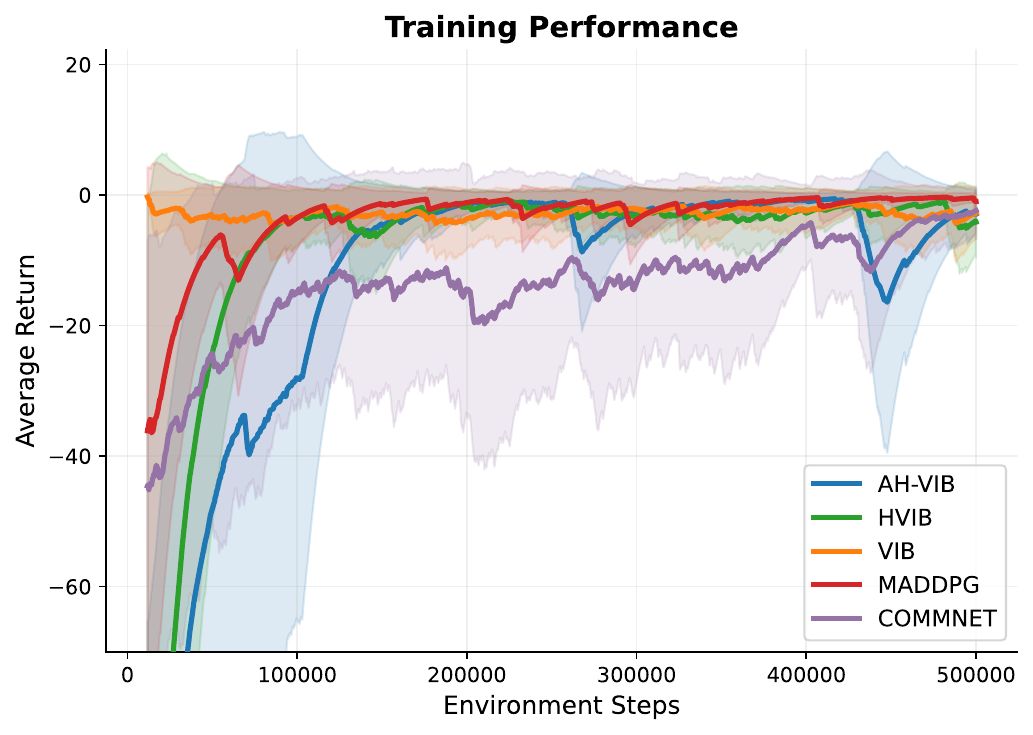}
    \caption{Mean episodic return over training steps for AH-VIB, HVIB, VIB,
    MADDPG, and CommNet, averaged across three seeds. Shaded regions denote
    $\pm 1$ standard deviation. All methods are trained under the same
    variable-bandwidths, with $\rho \sim \mathrm{Beta}(3.0,\,7.0)$.}
    \label{fig:training_reward}
\end{figure}
\subsection{Evaluation Metrics}
\begin{figure*}[!t]
    \vspace{2mm}
    \centering
    \includegraphics[width=\textwidth, trim=0cm 0.35cm 0cm 1.0cm, clip]
    {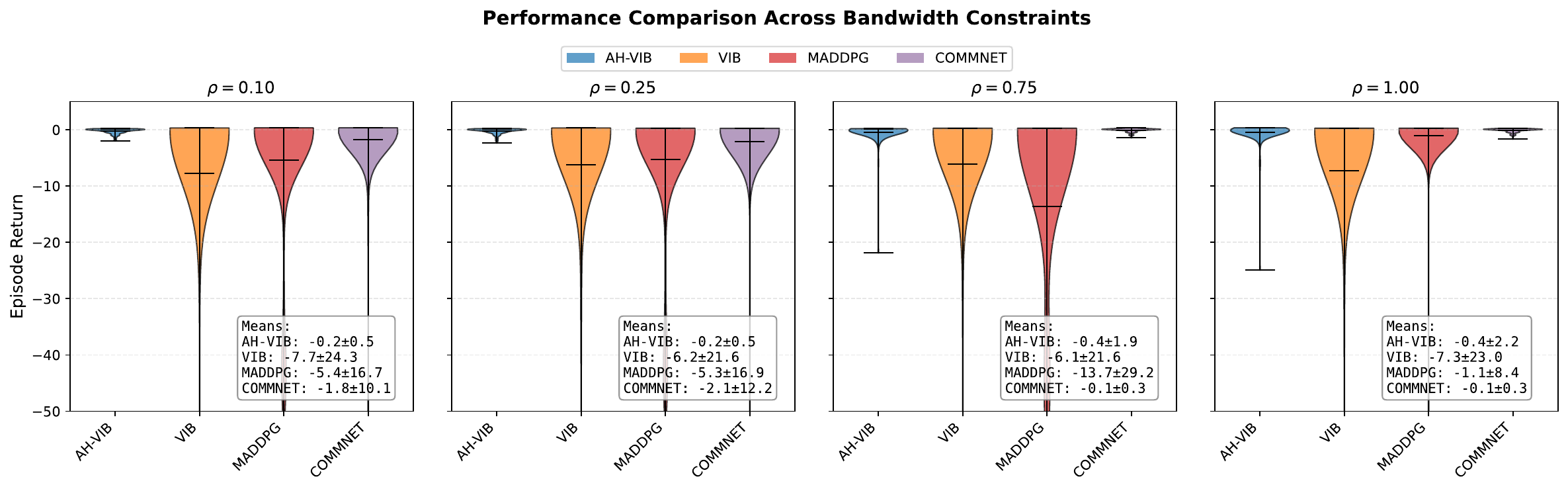}
    \caption{
    Episode return distributions at selected bandwidth fractions
    $\rho \in \{0.10,\,0.25,\,0.75,\,1.00\}$ for AH-VIB, VIB,
    MADDPG, and CommNet. Violin width indicates return density, and
    horizontal bars mark the mean.
    }
    \label{fig:violin}
\end{figure*}

\paragraph{Episode reward}
Our first metric is the mean undiscounted episode return $\bar{R}=\sum_{t=0}^{T-1} r^t$ , where higher (less negative) values indicate better performance. For training curves, returns are smoothed across episodes for visual clarity.

\paragraph{Robustness at constrained bandwidth}
At inference time, each trained policy is evaluated at fixed bandwidth fractions $\rho$, with truncation cutoff $k=\mathrm{round}(\rho d_m)$. This produces a reward--bandwidth curve that shows team performance when communication is restricted. 

\paragraph{Intra-episode bandwidth dynamics}
To test out-of-distribution robustness, we also evaluate policies under within-episode bandwidth schedules, where $\rho$ either decreases from $1.0$ to $0.0$ or increases from $0.0$ to $1.0$. Performance is measured by episode reward.

\begin{table}[b]
\centering
\setlength{\tabcolsep}{4pt}
\caption{
% Episode return at selected bandwidth fractions. Values show mean $\pm$ standard deviation. The best mean and best standard deviation per column are shown in \textbf{bold}, respectively.
Episode return at selected bandwidth fractions (mean $\pm$ standard deviation). Bold indicates the best mean and lowest standard deviation in each column.
}
\label{tab:ep_reward_slam}
\begin{tabular}{lllll}
\toprule
 & $\rho$=0.10 & $\rho$=0.25 & $\rho$=0.75 & $\rho$=1.0 \\
\midrule
AH-VIB & \textbf{-0.2} $\pm$ \textbf{0.5} & \textbf{-0.2} $\pm$ \textbf{0.5} & -0.4 $\pm$ 1.9 & -0.4 $\pm$ 2.2 \\
VIB & -7.7 $\pm$ 24.3 & -6.2 $\pm$ 21.6 & -6.1 $\pm$ 21.6 & -7.3 $\pm$ 23.0 \\
MADDPG & -5.4 $\pm$ 16.7 & -5.3 $\pm$ 16.9 & -13.7 $\pm$ 29.2 & -1.1 $\pm$ 8.4 \\
COMMNET & -1.8 $\pm$ 10.1 & -2.1 $\pm$ 12.2 & \textbf{-0.1} $\pm$ \textbf{0.3} & \textbf{-0.1} $\pm$ \textbf{0.3} \\
\bottomrule
\end{tabular}
\end{table}

\subsection{Training Configuration}

All algorithms share a common base training configuration. Each method is trained for $500,000$ environment steps with episode length $T=200$, replay buffer size $2 \times 10^5$, batch size $256$, and random exploration for the first $4{,}000$ steps. Updates are then performed every $50$ environment steps using $5$ gradient steps per update. Actor and critic learning rates are $3 \times 10^{-4}$ and $3 \times 10^{-5}$, respectively, with discount factor $\gamma=0.95$ and target-network update rate $\tau=0.005$.

For the VIB-based variants, we use bottleneck coefficient $\beta=0.001$. For the hierarchical variants, we use robustness-loss weight $\lambda=0.05$, a single sampled prefix per update, and a stop-gradient target on $Q_{\text{full}}$. The full hyperparameter details with algorithm-specific configurations are given in Table~\ref{tab:hyperparams}.

\subsection{Training Dynamics}
\label{sec:training}

We first examine whether the additional bottleneck and robustness terms in AH-VIB destabilize training or improve convergence. Figure~\ref{fig:training_reward} shows the mean episodic return over training steps, with shaded regions indicating one standard deviation.

AH-VIB, HVIB, VIB, and MADDPG converge to similar final returns, indicating that the VIB and hierarchical robustness objectives do not impede training. CommNet converges more slowly and remains more variable. Since asymptotic mean return only weakly separates the stronger methods, we focus on reliability under restricted bandwidth.

\paragraph{Statistical Reliability}
To formalize the observation of tighter return distributions, we perform bootstrap resampling ($N=10{,}000$) and compare the standard deviation of episode returns against the closest VIB-based baseline via $\Delta\sigma=\sigma_{\text{AH-VIB}}-\sigma_{\text{VIB}}$, where $\Delta\sigma<0$ indicates more consistent behavior. Across all evaluated bandwidth fractions $\rho\in\{0.10,0.25,0.75,1.00\}$, $\Delta\sigma$ is strongly negative (between $-19.7$ and $-23.8$) and the bootstrap probability that AH-VIB has lower standard deviation than VIB is $100\%$ at every bandwidth. The same holds against MADDPG ($P \geq 86.8\%$ at all bandwidths). Against CommNet, AH-VIB is significantly more consistent at the two tightest bandwidths ($P \geq 99.3\%$ at $\rho\in\{0.10,0.25\}$), while CommNet is the most consistent method at $\rho \geq 0.75$. This supports the concrete claim that AH-VIB's advantage under bandwidth constraints is improved reliability — and that this advantage is most pronounced precisely when communication is scarce.

\subsection{Performance Under Variable Bandwidth}
\label{sec:bw_performance}

We evaluate inference performance under fixed bandwidth budgets.
Figure~\ref{fig:violin} shows episode-return distributions at $\rho \in \{0.10,\,0.25,\,0.75,\,1.00\}$, and Table~\ref{tab:ep_reward_slam} reports the corresponding mean and standard
deviation.

AH-VIB achieves the highest mean return at the two tightest bandwidth settings, $\rho=0.10$ and $\rho=0.25$, where reliable use of a short message prefix is most important. At the larger bandwidth fractions, $\rho=0.75$ and $\rho=1.00$, CommNet obtains a slightly higher mean return and lower variation. This suggests that when most or all of the message is available, the additional structure imposed by AH-VIB is less important.

The principal advantage of AH-VIB is therefore most evident under restricted communication, often encountered in harsh environmental conditions. As shown in Figure~\ref{fig:violin}, AH-VIB produces a compact return distribution with short lower-return tails at $\rho=0.10$ and $\rho=0.25$. In contrast, VIB and MADDPG exhibit substantially broader distributions and more severe low-return episodes under these constrained conditions. CommNet remains competitive at larger bandwidth fractions, but does not provide the same mean-return performance as AH-VIB when only a small message prefix is received.

Table~\ref{tab:ep_reward_slam} shows the same bandwidth-dependent pattern. AH-VIB provides the best mean return and lowest return variation at the two tightest bandwidth settings, whereas CommNet performs best when bandwidth is less restrictive. Thus, AH-VIB is not intended to uniformly dominate every communication method at full bandwidth; rather, it improves the reliability of coordination when the available bandwidth is limited.

\subsection{Per-Dimension KL Divergence}
\label{sec:kl_divergence}

To better understand how the VIB-based methods use the information bottleneck, we analyse the per-dimension KL divergence between the encoder posterior and the unit-Gaussian prior:
\begin{equation}
D_{\mathrm{KL}}\!\left(\mathcal{N}(\mu_d,\sigma_d^2)\,\|\,\mathcal{N}(0,1)\right).
\end{equation}
The values are averaged over agents and timesteps within each evaluation episode, yielding one KL profile per episode.

We report this diagnostic analysis on MPE Simple Spread, which provides a controlled setting for comparing latent message usage across the VIB-based methods. All profiles are evaluated at full bandwidth ($\rho=1.0$), so they reflect native dimension usage rather than the direct effect of masking. While the main task-level evaluation in this paper uses the object-inspection and occupancy-mapping task, the KL analysis is included as complementary evidence about the learned communication representations.

\begin{figure}[tb]
    \centering
    \vspace{2mm}
    \includegraphics[
        width=\columnwidth,
        trim=0cm 0.3cm 0cm 0.24cm,
        clip
    ]{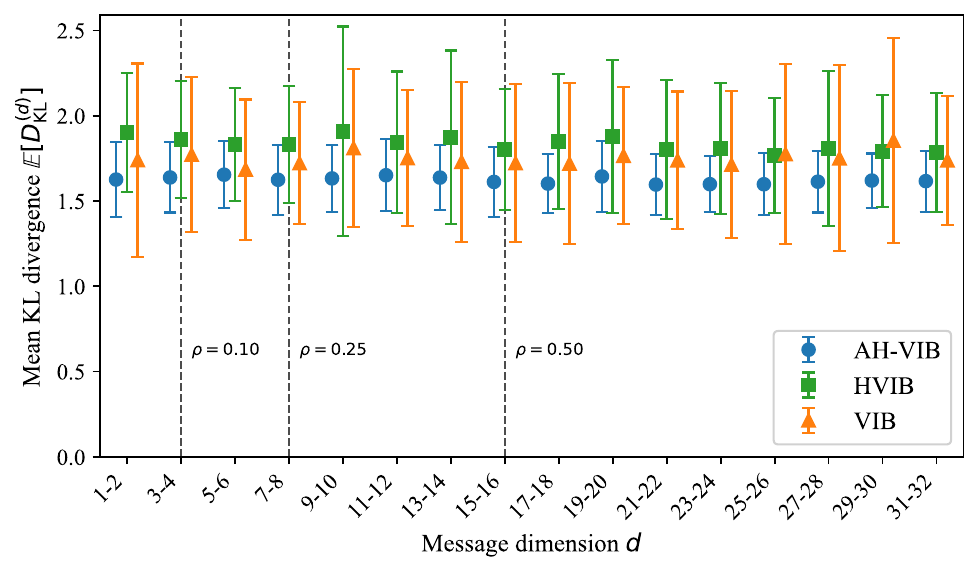}
    \caption{
    Mean per-dimension KL divergence between the encoder posterior and the unit-Gaussian prior for AH-VIB, HVIB, and VIB on MPE Simple Spread. Results are averaged over evaluation episodes at full bandwidth ($\rho=1.0$); error bars show the standard error of the mean (SEM). Vertical dashed lines indicate the prefix lengths corresponding to $\rho=0.10$, $\rho=0.25$, and $\rho=0.50$ for visual reference.
    }
    \label{fig:kl_profile}
\end{figure}

Figure~\ref{fig:kl_profile} shows that AH-VIB obtains lower mean KL divergence than HVIB and VIB across message dimensions, indicating a more compact latent communication representation~\cite{wang2020imac}. AH-VIB also exhibits tighter SEM bands, indicating more stable dimension-wise latent usage across evaluation episodes.

\subsection{Discussion and Future Work}\label{sec:limitations}

% While AH-VIB demonstrates superior performance under bandwidth constraints, limitations exist in this study. The communication setup is rather simple. 
While AH-VIB demonstrates superior performance under bandwidth constraints the simple communication setup presents a limitation.
Furthermore, agents broadcast messages at every timestep, no advanced communication media access protocol, and scheduling mechanisms and resource allocation are integrated. The communication channel model is also simplified without considering packet loss, propagation delay, and asymmetric links, which could have a significant impact for multi-agent collaboration, in particular, in underwater robot collaboration. 
In addition, the inspection environment remains a simplified simulation. It uses rectangular inspection objects, idealized ray-casting sensors, simplified motion dynamics, and perfect localization. These assumptions limit the conclusions that can be drawn about transfer to physical multi-robot inspection systems.

A natural direction for future work is to combine task-oriented communication with advanced scheduling and resource allocation, so that agents can learn both what information should appear early in a message and which agent can communicate at which slot. Another direction is to evaluate the method on more challenging tasks and more realistic multi-robot settings with practical channel modeling.

\section{Conclusion}

We introduced AH-VIB, a novel autoregressive variational communication method for multi-agent coordination under variable bandwidth and prefix truncation. AH-VIB combines causal message generation with a hierarchical robustness loss that promotes retention of task performance when only a message prefix is received.

On our cooperative object-inspection task, AH-VIB achieves competitive mean return while improving the consistency of performance under bandwidth restriction. These results support progressive message encoding as a useful design principle for bandwidth-constrained multi-robot communication.

\bibliographystyle{IEEEtran}
\bibliography{refs}

\end{document}